\documentclass[letterpaper]{article} 
\usepackage[submission]{aaai23}  
\usepackage{times}  
\usepackage{helvet}  
\usepackage{courier}  
\usepackage[hyphens]{url}  
\usepackage{graphicx} 
\usepackage{natbib}  
\usepackage{caption} 
\usepackage{algorithm}
\usepackage{algorithmic}

\usepackage{newfloat}
\usepackage{listings}
\usepackage{amssymb}
\usepackage{amsmath}
\usepackage{multirow}
\usepackage{amsthm}

\DeclareCaptionStyle{ruled}{labelfont=normalfont,labelsep=colon,strut=off} 
\floatstyle{ruled}
\newfloat{listing}{tb}{lst}{}
\floatname{listing}{Listing}
\title{Boosting Point-BERT by Multi-choice Tokens}
\author{
    Kexue Fu\textsuperscript{\rm 1}\textsuperscript{\rm 2}\equalcontrib,
    Mingzhi Yuan\textsuperscript{\rm 1}\textsuperscript{\rm 2}\equalcontrib,
    Manning Wang\textsuperscript{\rm 1}\textsuperscript{\rm 2}\thanks{Corresponding Author.}
}
\affiliations{
    \textsuperscript{\rm 1}Digital Medical Research Center, School of Basic Medical Science, Fudan University, Shanghai, China\\
    \textsuperscript{\rm 2}Shanghai Key Laboratory of Medical Image Computing and Computer Assisted Intervention, Shanghai, China

    kxfu18@fudan.edu.cn, mzyuan20@fudan.edu.cn, mnwang@fudan.edu.cn
}

\usepackage{bibentry}

\begin{document}

\maketitle

\begin{abstract}
    Masked language modeling (MLM) has become one of the most successful self-supervised pre-training task. Inspired by its success, Point-BERT, as a pioneer work in point cloud, proposed masked point modeling (MPM) to pre-train point transformer on large scale unanotated dataset. Despite its great performance, we find the inherent difference between language and point cloud tends to cause ambiguous tokenization for point cloud. For point cloud, there doesn't exist a gold standard for point cloud tokenization. Point-BERT use a discrete Variational AutoEncoder (dVAE) as tokenizer, but it might generate different token ids for semantically-similar patches and generate the same token ids for semantically-dissimilar patches. To tackle above problem, we propose our McP-BERT, a pre-training framework with multi-choice tokens. Specifically, we ease the previous single-choice constraint on patch token ids in Point-BERT, and provide multi-choice token ids for each patch as supervision. Moreover, we utilitze the high-level semantics learned by transformer to further refine our supervision signals. Extensive experiments on point cloud classification, few-shot classification and part segmentation tasks demonstrate the superiority of our method, e.g., the pre-trained transformer achieves 94.1\% accuracy on ModelNet40, 84.28\% accuracy on the hardest setting of ScanObjectNN and new state-of-the-art performance on few-shot learning. We also demonstrate that our method not only improves the performance of Point-BERT on all downstream tasks, but also incurs almost no extra computational overhead. The code will be released in \url{https://github.com/fukexue/McP-BERT}.
\end{abstract}

\section{Introduction}

Self-supervised pre-training \cite{ref1,ref2,ref3,ref4,ref5,ref6,ref7,ref8} is attracting growing attentions as it can transfer knowledge learned from large scale unlabeled dataset to boost the performance on downstream tasks. 
Most self-supervised pre-training methods are based on specific proxy tasks such as permutation prediction \cite{ref6}, image colorization \cite{ref7}, instance-level discrimination \cite{ref1,ref2,ref8}. 
Among them, the masked language modeling (MLM) task proposed in BERT \cite{ref9} is currently one of the most successful proxy tasks and has been migrated to many other domains \cite{ref10,ref11,ref12}. 
Point-BERT \cite{ref12}, as a pioneer work in point cloud learning, proposed a variant of MLM called masked point modeling (MPM) to pre-train point cloud transformers. 
Specifically, it first divides a point cloud into several local point patches and assigns a token id to each patch to convert the point cloud into multiply discrete tokens. 
Then it masks a proportion of tokens and pre-trains the model by recovering the masked tokens based on the transformer encoding results of the corrupted point cloud. 
Since there is not well-defined vocabulary to generate token ids for local point cloud patches, Point-BERT utilizes a pre-trained discrete Variational AutoEncoder (dVAE) \cite{ref13} as the tokenizer. 
However, we find such discrete tokenization in MPM hinders the framework to achieve better performance due to the difference between point cloud and language. 

\begin{figure}[tp]
    \centering
    \includegraphics[width=0.46\textwidth]{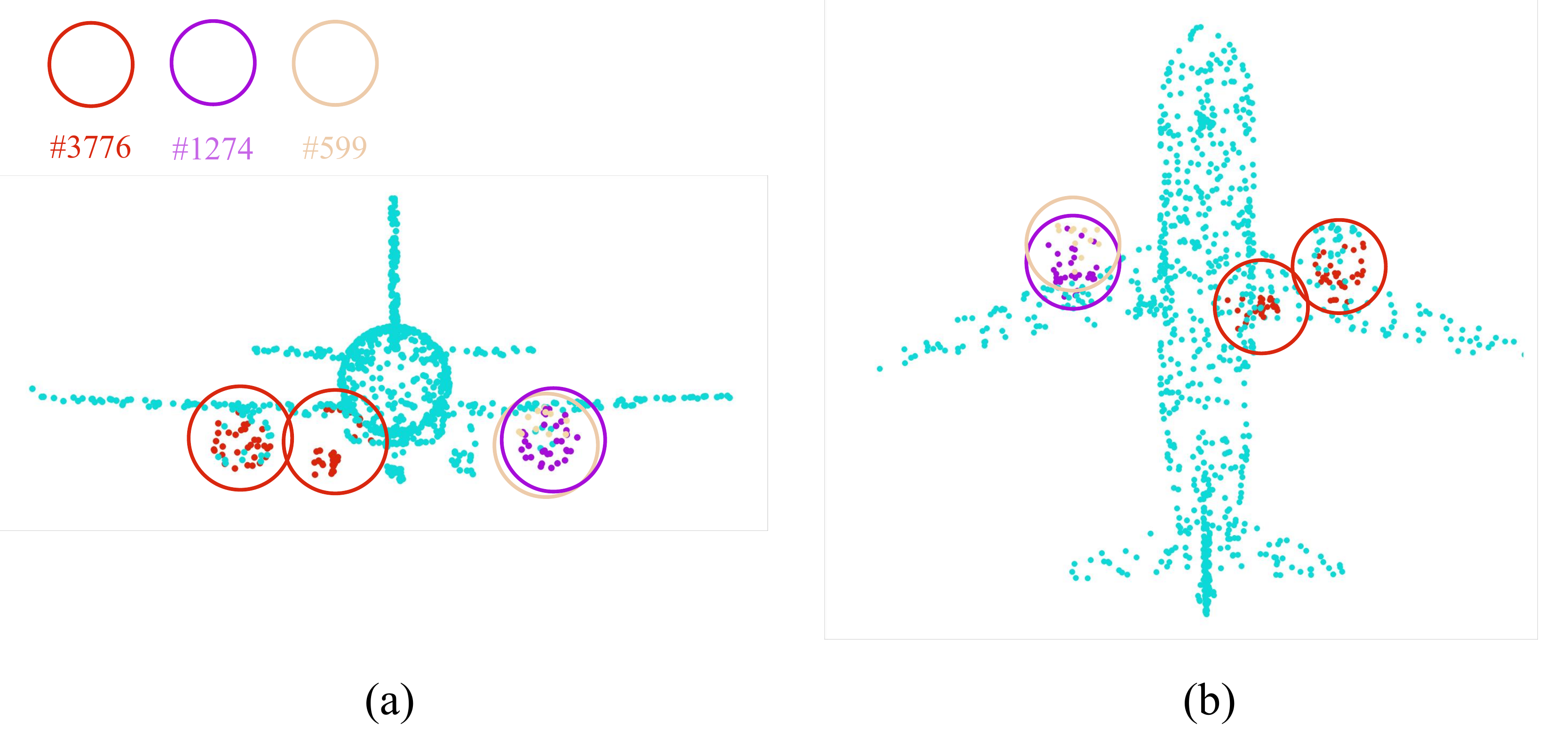}
    \caption{
        \textbf{Visualization of improper token ids.} 
        To better visualize the improper tokenization, we provide two different views of the same point cloud. 
        Different colors represent the different token ids, and the circles indicate the range of local patches. 
        As shown in above two figures, semantically-dissimilar patches (landing gear and aero-engine, colored in red) have the same token ids (\#3776), and semantically-similar adjacent patches (aero-engine, colored in purple and brown) have different token ids (\#1274, \#599). 
    }
    \label{fig1}
\end{figure}

Languages are naturally composed of discrete words, which are strictly bijective to the token ids. 
However, there is no such gold standard for point clouds discretization as for language, making it inevitable to introduce noise to MPM. 
Specifically, dVAE-based tokenization tends to cause the following two kinds of ambiguities. 
(1) Semantically-dissimilar patches have the same token ids. 
Due to the lack of well-defined vocabulary, MPM adopts a pre-trained dVAE as tokenizer to generate token ids for local patches. 
However, the tokenizer focuses on local patches' geometry but barely considers their semantics, resulting in some wrong token ids. 
For example, two patches shown in red in Figure \ref{fig1} consist of similar geometric structure but have different semantics (landing gear and aero-engine). 
However, they are allocated with the same token ids (\#3776) due to their similar geometric structure. 
(2) Semantically-similar patches have different token ids. 
As shown in Figure \ref{fig1}, semantically-similar patches of the aero-engine of the airplane are allocated with many different token ids (\#599, \#1274). 
It's reasonable for them to have the same token ids because of their similar semantics and geometry to each other. 
However, the tokenizer neglects their relations and allocates them with different token ids due to the interference of imperfect discretization and acquisition noise. 

Inspired by mc-BEiT \cite{ref11}, we propose an improved BERT-style point cloud pre-training framework called McP-BERT with multi-choice tokens to tackle the above problems. Specifically, since semantically-similar patches might have different token ids, we ease the previous strict single-choice constraint on patch token ids. For each local patch, we utilize a probability distribution vector of token ids as supervision signal rather than previous single token id, which means each patch has \textbf{M}ulti-\textbf{C}hoice token ids. 
Moreover, we believe high-level perceptions produced by point cloud transformer can provide extra semantic supervision signals, which benefit our pre-training. 
Therefore, we refine the above supervision signals, i.e., the probability distribution vectors, using inter-patch semantic similarities, which come from our point cloud transformer. 
By considering high-level similarities, the ambiguities caused by local geometric similarity can be mitigated.

To verify the effectiveness of our framework, we pre-train a point cloud transformer on ShapeNet \cite{ref14} and conduct extensive fine-tuning experiments on downstream tasks including point cloud classification, few-shot classification and part segmentation. 
Our framework not only improve the performance of previous Point-BERT on \textbf{all} downstream tasks, but also incurs almost no extra computational overhead during pre-training. 
Our McP-BERT achieves 94.1\% accuracy on ModelNet40 \cite{ref24} and 84.28\% accuracy on the complicated setting of ScanObjectNN \cite{ref47}, outperforming a series of state-of-the-art methods. Our method also achieves new state-of-the-art on point cloud few-shot learning, indicating powerful generalization of our McP-BERT. 

\section{Related Works}
Pre-training on large-scale data and then fine-tuning on target tasks has been proved to be an effective paradigm to boost the performance of model on downsteam tasks \cite{ref15}. However, although many efficient annotation tools \cite{ref16, ref17} have been proposed, labeling a large-scale dataset is still costly especially for point cloud data. Therefore, self-supervised pre-training which pre-trains models without annotations has more potential. In the past few years, many self-supervised pre-training methods for point cloud learning had been proposed and the core of them is designing a proper proxy task. For example, Poursaeed et al. \cite{ref34} pre-trained networks by predicting point cloud’s rotation. Inspired by the success of a series of contrastive learning-based self-supervised method \cite{ref1, ref2, ref8}, PointContrast \cite{ref36} and STRL \cite{ref35} implemented a contrastive learning paradigm to learn deep representations from depth scan. OcCo \cite{ref37} pre-trained their encoder by reconstructing occluded point clouds. Recently, inspired great success of BERT \cite{ref9}, Point-BERT \cite{ref12} designed a proxy task called masked point modeling (MPM), following a mask-and-then-predict paradigm to pre-train point cloud transformer. Since there exist no vocabulary for officially break raw point clouds into small predefined chunks, Point-BERT incorporates a dVAE as tokenizer. It first divides a complete point cloud into several patches and then allocates token ids for patches based on dVAE’s latent features. However, we find that as we analyzed above, this improper tokenizer tends to produce ambiguous token ids for each patch. Therefore, we improve Point-BERT by introducing eased and refined token ids in this work. 

\section{Preliminaries}
\begin{figure*}[h]
    \centering
    \includegraphics[width=0.995\textwidth]{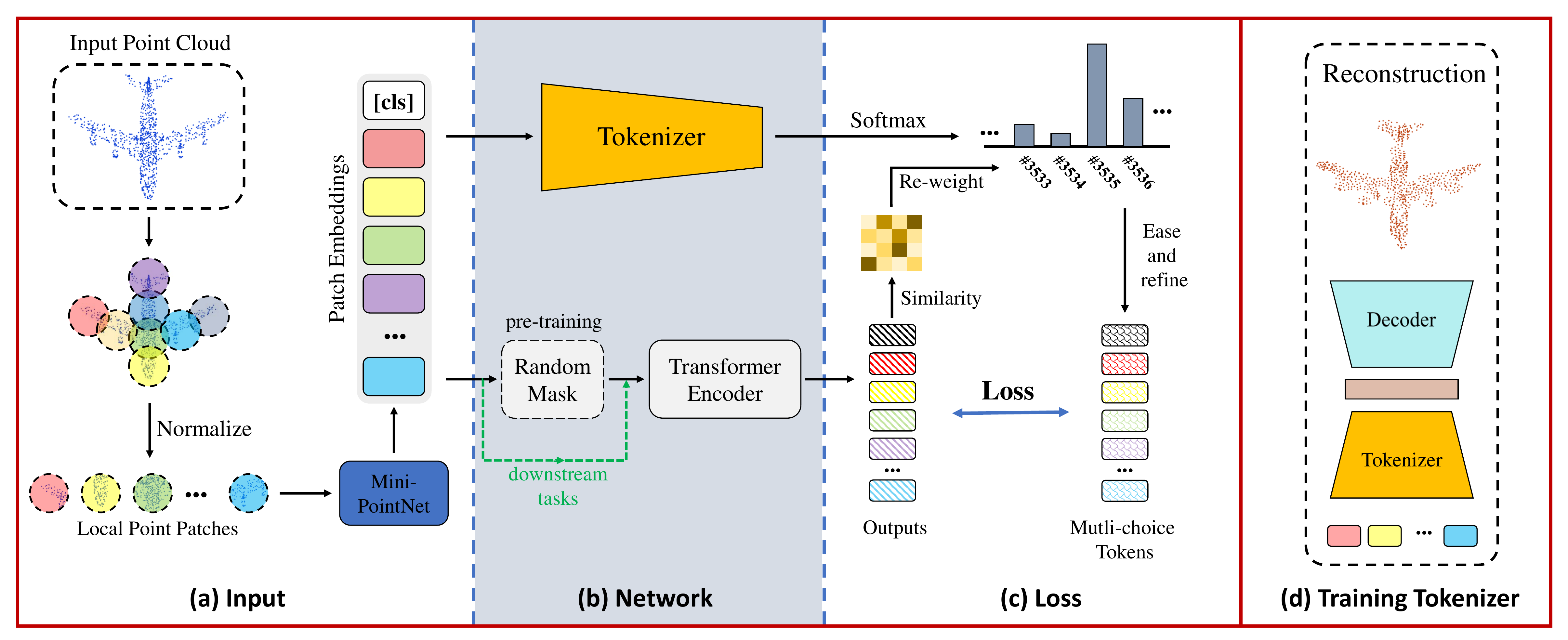}
    \caption{
        \textbf{Overview of our proposed method, McP-BERT.} 
        We improve the Point-BERT by incorporating multi-choice tokens. We utilize a softmax layer to ease the hard token ids into soft probability distribution vectors of token ids and use the encoded features generated by transformer encoder to re-weight the soft probability distribution vectors. During pre-training, a proportion of tokens are randomly masked and fed into transformer encoder. The transformer encoder is optimized by predicting the masked eased and refined token signals in the form of a soft-label cross-entropy loss. Unlike pre-training, when applied to downstream tasks, we skip the random mask operation, which is represented by a dashed green line, and input all patch embeddings into the transformer encoder.
    }
    \label{fig2}
\end{figure*}

Similar to Point-BERT \cite{ref12}, we also adopt a Masked Point Modeling (MPM) paradigm as our proxy task to pre-train our model. 
Specifically, given a point cloud $P$, we first sample $g$ center points $\left\{c_{i}\right\}_{i=1}^{g}$ using farthest point sampling \cite{ref38} and then select their $k$-nearest neighbor points to build $g$ local patches $\left\{p_{i}\right\}_{i=1}^{g}$. 
The points in local patches are normalized by subtracting their center points to further mitigate their coordinate bias. 
Then a mini-PointNet \cite{ref25} is used to map the normalized patches to a sequence of patch embeddings $\left\{f_{i}\right\}_{i=1}^{g}$. The above process is shown in Figure \ref{fig2}(a). After that, a tokenizer will take embeddings as inputs and generate token ids $\left\{\hat{z}_{i} \in \mathbb{R}^{|V|}\right\}_{i=1}^{g}$ for patches, where ${|V|}$ denotes the length of the vocabulary $V$. 
As done in other BERT-style works, we randomly mask a proportion of tokens and then send corrupted patch embeddings into a backbone implemented by transformer encoder. The backbone will learn $l2$-normalized representations $\left\{h_{i}\right\}_{i=1}^{g}$ for both masked and unmasked tokens, as shown in Figure \ref{fig2}(b). Finally, we pre-train the backbone by predicting the masked tokens based on these representations. The objective of MPM can be formulated as follow:
\begin{equation}
    \mathcal{L}_{M P M}=\mathbb{E}_{i \in M}\left[-\sum_{k=1}^{|V|} \hat{z}_{i, k} \log q\left(\hat{z}_{i, k} \mid h_{i}\right)\right]
\end{equation}
where $M$ denotes the set of masked patches, $q$ denotes a MLP head to predict the token id $\hat{z}_{i}$ of masked patch $p_i$ based on representation $h_i$. 
For Point-BERT, the token id $\hat{z}_{i}$ is a one-hot vector, which is derived from latent feature $z_i$ in dVAE:
\begin{equation}
    \hat{z}_{i, k}=\left\{\begin{array}{lr}
    1 & \text { if } k=\underset{j}{\arg \max } z_{i, j} \\
    0 & \text { else }
    \end{array}\right.
\end{equation}
where $z_{i} \in \mathbb{R}^{|V|}$ denotes latent feature in the dVAE's encoder. 
However, as mentioned above and shown in Figure \ref{fig1}, this imperfect tokenizer will make some semantically-similar patches have different tokens and some semantically-dissimilar patches have the same tokens. To tackle these problems, we ease the token id $\hat{z}_{i}$ to a soft vector satisfying $\sum_{k} \hat{z}_{i, k}=1$, allowing one patch corresponds to multiply token ids. Moreover, we re-build $\hat{z}_{i}$ based on the high-level semantic relationship for more accurate supervision. The details are presented in the next section.

\section{Method}
Our framework is shown in Figure \ref{fig2}, and we will introduce it in detail in this section. 
To verify the effectiveness of our proposed idea, we set Point-BERT as our baseline and modify it as little as possible. 

Before pre-training, we train a dVAE \cite{ref13} as our tokenizer in the same way as in Point-BERT. 
During pre-training, transformer \cite{ref30} encoder in our framework is trained on unlabeled data over MPM. 
For fine-tuning, the transformer encoder is first initialized with the pre-trained parameters, and all of the parameters are fine-tuned using labeled data from downstream tasks. 
We first briefly introduce the implements of the tokenizer and the transformer encoder. Then we provide details about our multi-choice strategy, which plays an important role in improving the tokenization. 

\subsection{Tokenizer}\label{sec41}
Our tokenizer is trained through dVAE-based point cloud reconstruction. 
As shown in Figure \ref{fig2}(d), the dVAE adopts an encoder-decoder architecture. 
The encoder is our tokenizer $Q_{\phi}\left(z^{o} \mid p\right)$, which consists of a DGCNN \cite{ref27} and Gumbel-softmax \cite{ref40}. 
During training dVAE, DGCNN $Q_{\phi}\left(z \mid p\right)$ takes a sequence of patch embeddings as input and outputs latent features $\left\{z_{i}\right\}_{i=1}^{g}$. 
Then the latent features $\left\{z_{i}\right\}_{i=1}^{g}$ is discretized into one-hot vectors $\left\{z_{i}^{o}\right\}_{i=1}^{g}$ through a Gumbel-softmax. 
The decoder $P_{\theta}\left(p \mid z^{o}\right)$ consists of a DGCNN followed by a FoldingNet \cite{ref39}. 
DGCNN takes one-hot vectors $\left\{z_{i}^{o}\right\}_{i=1}^{g}$ as input and take full advantage of their neighborhood relationship in the feature space to enhance the representation of these discrete tokens. 
The following FoldingNet reconstructs the original point cloud according to DGCNN's outputs. 
As with other VAE-style works \cite{ref41}, we achieve reconstruction objective by optimizing the evidence lower bound (ELBO) of the log-likelihood $P_{\theta}(p \mid \tilde{p})$:
\begin{equation}
    \begin{aligned}
    \sum_{\left(p_{i}, \tilde{p}_{i}\right) \in D} & \log P_{\theta}\left(p_{i} \mid \widetilde{p}_{i}\right) \geq \\  \sum_{\left(p_{i}, \tilde{p}_{i}\right) \in D}&\left(\mathbb{E}_{z_{i}^{o} \sim Q_{\phi}\left(z^{o} \mid p_{i}\right)}\left[\log P_{\theta}\left(p_{i} \mid z_{i}^{o}\right)\right]\right.\\
    &\left.-K L\left[Q_{\phi}\left(z^{o} \mid p_{i}\right) \| P_{\theta}\left(z^{o} \mid \tilde{p}_{i}\right)\right]\right)
    \end{aligned}
\end{equation}
where $\widetilde{p}$ denotes the reconstructed patches, $D$ denotes the training corpus, $P_{\theta}\left(z^{o} \mid \tilde{p}_{i}\right)$ follows a uniform distribution. 
The former of ELBO represents the reconstruction loss and the latter represents the distribution loss. 
We follow \cite{ref42} to calculate the reconstruction loss and follow \cite{ref43} to optimize the distribution difference between one-hot vectors and prior. 
Our tokenizer has the same architecture as in Point-BERT \cite{ref12} but outputs the latent features $\left\{z_{i}\right\}_{i=1}^{g}$ rather than the one-hot vectors $\left\{z_{i}^{o}\right\}_{i=1}^{g}$ during pre-training. 
More details about the architecture of tokenizer can refer to Point-BERT.

\subsection{Transformer Encoder}\label{sec42}
We utilize a standard transformer \cite{ref30} following Point-BERT’s settings as our backbone, as shown in Figure \ref{fig2}(b). During pre-training, we randomly mask a proportion of patch embeddings and fed them including masked and unmasked patch embeddings into the backbone. Then the backbone outputs the deep representations $\left\{h_{c l s}\right\} \cup\left\{h_{i}\right\}_{i=1}^{g}$, where $h_{i}$ represents the representations for both masked and unmasked patches and $h_{cls}$ represents the representation for classification, as part of the input of classification head in the downstream task. The deep representations extracted by transformer consider the associations between patches and contain more semantic information, which is suitable for many downstream tasks and refining the supervision signals. We will go over how to use these representations to refine the supervision signals in the following section. 


\subsection{Multi-choice Tokenization}\label{sec43}
As mentioned above, there doesn't exist a gold standard for point cloud tokenization. 
Therefore, it's inevitable for tokenizer to produce improper supervision, including generating the same token ids for semantically-dissimilar patches and generating different token ids for semantically-similar patches. 
We observe that given a local patch, there may exist multiply suitable token ids as candidates as illustrated in Figure \ref{fig1}. 
Inspired by mc-BEiT \cite{ref11}, we attempt to release the strict single-choice constraint on token ids. 
As shown in Figure \ref{fig2}(c), given a local patch $p_i$, we don't use unique token id as supervision any more. Instead, we predict the probability distribution vector $\mathcal{P}\left(z_{i}\right) \in \mathbb{R}^{|V|}$ of the token ids. Specifically, the probability distribution vector $\mathcal{P}\left(z_{i}\right) \in \mathbb{R}^{|V|}$ is generated by a softmax operation: 
\begin{equation}
    \mathcal{P}\left(z_{i}\right)_{k}=\frac{\exp \left(z_{i, k} / \tau\right)}{\sum_{j=1}^{|V|} \exp \left(z_{i, j} / \tau\right)}
\end{equation}
where $z_{i} \in \mathbb{R}^{|V|}$ denotes the latent feature in tokenizer, $\tau$ is a temperature coefficient, which controls the smoothness of probability distribution. When $\tau$ is small, $\mathcal{P}\left(z_{i}\right)$ tends to be a one-hot vector. When $\tau$ is large, it is equivalent to incorporates more choices. 

Moreover, to further tackle the ambiguity of token ids as shown in Figure \ref{fig1}, more high-level semantics should be incorporated. As analyzed in introduction, tokenizer focuses on encoding local geometry of local patches while ignoring the associations between patches, making semantically-dissimilar patches have the same token ids. To alleviate this problem, we use the representations learned by transformer to re-weight the probability distribution vectors as done in \cite{ref11}. 
Specifically, as shown in Figure \ref{fig2}(c), we use the cosine similarity between patch features learned by transformer to re-weight the probability distribution matrix $\mathcal{P}(z) \in \mathbb{R}^{g \times|V|}$. 
The similarity matrix $W \in \mathbb{R}^{g \times g}$ is calculated as follows: 
\begin{equation}
    W_{i, k}=\frac{\exp <h_{i}, h_{k}>}{\sum_{j=1}^{g} \exp <h_{i}, h_{j}>}
\end{equation}
where $h_i$ denotes the $l2$-normalized representations learned by transformer, $<\cdot,\cdot>$ denotes the inner product between two vectors. 
The refined probability distribution matrix $W\mathcal{P}(z)$ considers the inter-patch associations, making semantically-similar patches have more similar probability distributions and semantically-dissimilar patches have more discriminable probability distributions. 

The final targets $\hat{z}$ for prediction is a weighted sum of the eased probability distribution matrix $\mathcal{P}(z)$ and the refined probability distribution matrix $W \mathcal{P}(z)$:
\begin{equation}
    \hat{z}=\omega \mathcal{P}(z)+(1-\omega) W \mathcal{P}(z)
\end{equation}
where $\omega$ is a coefficient balancing the low-level semantics from tokenizer and high-level semantics from inter-patch similarity. 
Slightly different from mc-BEiT \cite{ref11}, we adopt a warm-up strategy, i.e. gradually decreasing $\omega$ rather than a constant one. Specifically, at the beginning of pre-training, we set $\omega=1$ due to inadequate training of transformer. 
After 30 epochs, transformer can well encode both the local geometry of local patches and dependencies between patches. 
Then the coefficient $\omega$ begins to decrease in a cosine schedule to boost pre-training. 
We set the lower bound of $\omega$ as 0.8. 
Our experiments also show this gradually decreasing paradigm plays an essential role in the pre-training. 
When we set $\omega$ to a constant at the beginning, the transformer is easy to collapse due to the noisy semantics from initial transformer. 

\section{Experiment}
In this section, we first introduce the setup of pre-training. 
Then we conduct experiments on downstream tasks including point cloud classification, few-shot learning and part segmentation. 
We also present various ablations to analyze the effectiveness of our method and provide visualizations of our learned representations. 
Furthermore, we also compare the computational overhead of our McP-BERT with Point-BERT. \textbf{Our code will be publicly available.}

\subsection{Pre-training}\label{sec51}
For all experiments, we use ShapeNet \cite{ref14} as the pre-training dataset, which originally cotains over 50,000 CAD models from 55 common object categories. We randomly sample 1024 points from each CAD model and divide them into 64 local patches. Each patch contains 32 points. For the BERT-style pre-taining, we randomly mask $25\sim45\%$ patches for MPM in a block masking manner \cite{ref12}. For more details of implementation, please refer to the supplementary materials.

\begin{table}[]
    \centering
    \caption{ 
        \textbf{Comparisons of the classification on ModelNet40} \cite{ref24}. * represents training from scratch.
    }
    \setlength{\tabcolsep}{0.5cm}
    {
    \begin{tabular}{l r}
    \hline
    Method                    & Acc (\%)      \\ \hline
    PointNet \cite{ref25}         & 89.2          \\ 
    PointNet++ \cite{ref26}       & 90.5          \\ 
    PointCNN \cite{ref48}        & 92.2          \\ 
    DGCNN \cite{ref27}            & 92.2          \\ 
    DensePoint \cite{ref50}       & 92.8          \\ 
    KPConv \cite{ref28}           & 92.9          \\ 
    PTC \cite{ref33}              & 93.2          \\ 
    PointTransformer \cite{ref31} & 93.7          \\ 
    GLR \cite{ref49}             & 92.9          \\ 
    SRTL \cite{ref35}           & 93.1          \\ 
    Transformer*              & 91.4          \\ 
    OcCo \cite{ref37}            & 92.2          \\ 
    Point-BERT \cite{ref12}      & 93.8          \\ 
    Ours          & \textbf{94.1} \\ \hline
    \end{tabular}
    }
    \label{table1}
\end{table}

\begin{figure}[th]
    \centering
    \includegraphics[width=3in]{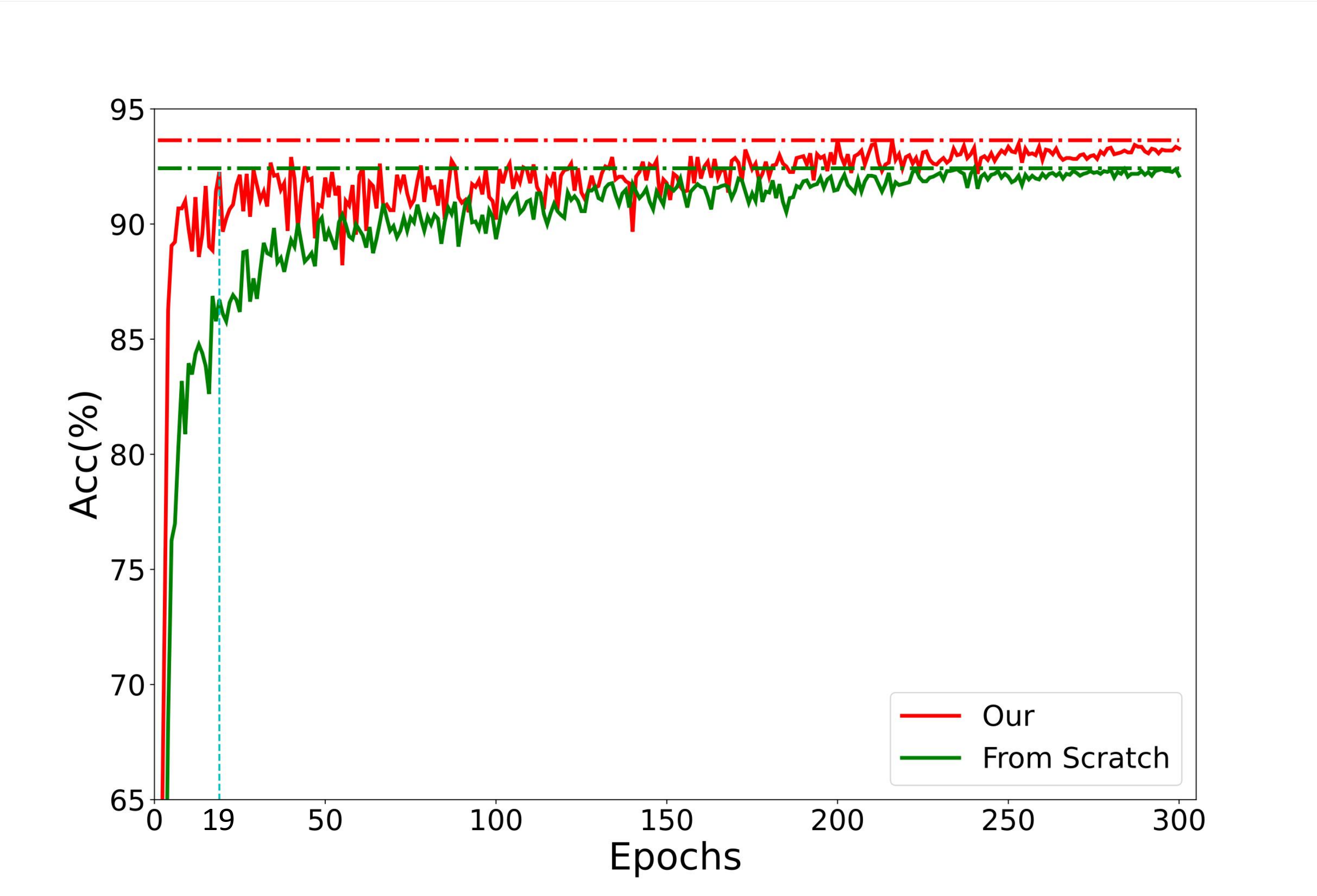}
    \caption{
        \textbf{Convergence curve.} 
        We compare the performance of transformers training from scratch (green) and pre-training with our method (red) in terms of validation accuracy (\%) on ModelNet40. 
        The red dotted line denotes the best performance of our method while the green dotted line denotes the best performance of the baseline. 
        Our method reaches baseline's best performance in only 19 epochs.
    }
    \label{fig3}
\end{figure}

\subsection{Point Cloud Classification}\label{sec52}
We perform our classification experiments on both the synthetic dataset ModelNet40 \cite{ref24} and real-world dataset ScanObjectNN \cite{ref47}.

\subsubsection{Experiment on Synthetic Dataset:} ModelNet40 \cite{ref24} is the most popular 3D dataset for point cloud classification, which contains 12311 CAD models from 40 object categories. 
We randomly sample 8k points from each CAD model for training and test. 
We follow previous setting \cite{ref25,ref16} to split the dataset into 9843/2468 for training and test.   

\noindent
\textbf{Competitors:} We compare our method with supervised methods, i.e., PointNet \cite{ref25}, PointNet++ \cite{ref26}, PointCNN \cite{ref48}, DGCNN \cite{ref27}, DensePoint \cite{ref50}, KPConv \cite{ref28}, PCT \cite{ref33}, PointTransformer \cite{ref31} and recently published self-supervised pre-training methods \cite{ref12,ref35,ref37,ref49}. 
Moreover, to illustrate the effectiveness of pre-training, we also set a baseline model, which use the same backbone model as ours but trained from scratch.

\noindent
\textbf{Results:} We adopt the classification accuracy as the evaluation metric and the experiment results are listed in Table \ref{table1}. 
As observed from the results, our method obtain 94.1\% accuracy on ModelNet40, outperforming the competitive methods and achieving new state-of-the-art performance. 
We also provide a convergence curve in  Figure \ref{fig3}. 
We can see our method improves the baseline by 2.7\% and converges to a high performance faster. 
Our method also outperforms Point-BERT by 0.3\% with negligible extra computational overhead, which indicates the effectiveness of our multi-choices strategy.

\begin{table}[]
    \centering
    \caption{ 
        \textbf{Comparisons of the classification on ScanObjectNN} \cite{ref47}. 
        We report the accuracy (\%) of three different settings (OBJ-BG, OBJ-ONLY, PB-T50-RS). * represents training from scratch.
    }
    {
    \begin{tabular}{ l|c c c }
    \hline
    Method                  & OBJ-BG         & OBJ-ONLY       & PB-T50-RS      \\ \hline
    PointNet        & 73.3           & 79.2           & 68.0           \\ 
    SpiderCNN       & 77.1           & 79.5           & 73.7           \\ 
    PointNet++      & 82.3           & 84.3           & 77.9           \\ 
    PointCNN       & 86.1           & 85.5           & 78.5           \\ 
    DGCNN          & 82.8           & 86.2           & 78.1           \\ 
    BGA-DGCNN       & -              & -              & 79.7           \\ 
    BGA-PN++  & -              & -              & 80.2           \\ \hline
    Transformer*                & 79.86          & 80.55          & 77.24          \\ 
    OcCo            & 84.85          & 85.54          & 78.79          \\ 
    Point-BERT     & 87.43          & 88.12          & 83.07          \\ 
    Ours                    & \textbf{88.98} & \textbf{90.02} & \textbf{84.28} \\ \hline
    \end{tabular}
    }
    \label{table2}
\end{table}

\subsubsection{Experiment on Real-world Dataset:} To test our method's generalization to real scenes, we also conduct an experiment on a real-world dataset. ScanObjectNN \cite{ref47} is a dataset modified from scene mesh datasets SceneNN \cite{ref52} and ScanNet \cite{ref53}. 
It contains 2902 point clouds from 15 categories. 
We follow previous works \cite{ref12,ref47} to carry out experiments on three variants: OBJ-BG, OBJ-ONLY and PB-T50-RS, which denote the version with background, the version without background and the version with random perturbations, respectively. 
More details can be found in \cite{ref47}. 

\noindent
\textbf{Competitors:} We compare our method with supervised methods, i.e., PointNet \cite{ref25}, SpiderCNN \cite{ref51}, PointNet++ \cite{ref26}, PointCNN \cite{ref48}, DGCNN \cite{ref27}, models equipped with background-aware \cite{ref47} (BGA) module and some state-of-the-art self-supervised pre-training methods \cite{ref12,ref37}. 
In this experiment, we also set a baseline as we do on ModelNet40. 

\noindent
\textbf{Results:} The classification accuracy on ScanObjectNN is shown in Table \ref{table2}. 
As observed from the results, all the methods perform worse on real-world dataset than on synthetic dataset ModelNet40, which is caused by less data for fine-tuning and the interference of noise, background, occlusion, etc. 
However, our method still achieves best performances on all three variants. 
It's worth noting that our method significantly improves the baseline by 9.12\%, 9.47\%, and 7.04\% on the three variants, which strongly confirms the generalization of our method. 
The above results also indicate that pre-training can transfer useful knowledge to downstream task and plays important roles in downstream task especially when downstream task is challenging.

\begin{table}[]
    \centering
    \caption{
        \textbf{Few-shot classification results on ModelNet40} \cite{ref24}. 
        We list the average accuracy (\%) and the standard deviation over 10 independent experiments. * represents training from scratch.
    }
    \setlength{\tabcolsep}{1.5mm}
    {
    \begin{tabular}{lcc|cc}
    \hline
    \multirow{2}{*}{}        & \multicolumn{2}{c|}{5-way}                                 & \multicolumn{2}{c}{10-way}                                \\ \cline{2-5} 
                             & \multicolumn{1}{c|}{10-shot}           & 20-shot           & \multicolumn{1}{c|}{10-shot}           & 20-shot           \\ \hline
    3D GAN          & \multicolumn{1}{|c|}{55.8±3.4}          & 65.8±3.1          & \multicolumn{1}{c|}{40.3±2.1}          & 48.4±1.8          \\ 
    FoldingNet     & \multicolumn{1}{|c|}{33.4±4.1}          & 35.8±5.8          & \multicolumn{1}{c|}{18.6±1.8}          & 15.4±2.2          \\ 
    L-GAN      & \multicolumn{1}{|c|}{41.6±5.3}          & 46.2±6.2          & \multicolumn{1}{c|}{32.9±2.9}          & 25.5±3.2          \\ 
    3D-Caps & \multicolumn{1}{|c|}{42.3±5.5}          & 53.0±5.9          & \multicolumn{1}{c|}{38.0±4.5}          & 27.2±4.7          \\ 
    PointNet        & \multicolumn{1}{|c|}{52.0±3.8}          & 57.8±4.9          & \multicolumn{1}{c|}{46.6±4.3}          & 35.2±4.8          \\ 
    PointNet++     & \multicolumn{1}{|c|}{38.5±4.4}          & 42.4±4.5          & \multicolumn{1}{c|}{23.1±2.2}          & 18.8±1.7          \\ 
    PointCNN        & \multicolumn{1}{|c|}{65.4±2.8}          & 68.6±2.2          & \multicolumn{1}{c|}{46.6±1.5}          & 50.0±2.3          \\ 
    RSCNN          & \multicolumn{1}{|c|}{65.4±8.9}          & 68.6±7.0          & \multicolumn{1}{c|}{46.6±4.8}          & 50.0±7.2          \\ 
    DGCNN          & \multicolumn{1}{|c|}{31.6±2.8}          & 40.8±4.6          & \multicolumn{1}{c|}{19.9±2.1}          & 16.9±1.5          \\ \hline
    Transformer*                 & \multicolumn{1}{|c|}{87.8±5.2}          & 93.3±4.3          & \multicolumn{1}{c|}{84.6±5.5}          & 89.4±6.3          \\ 
    OcCo           & \multicolumn{1}{|c|}{94.0±3.6}          & 95.9±2.3          & \multicolumn{1}{c|}{89.4±5.1}          & 92.4±4.6          \\ 
    Point-BERT     & \multicolumn{1}{|c|}{94.6±3.1}          & 96.3±2.7          & \multicolumn{1}{c|}{91.0±5.4}          & 92.7±5.1          \\ 
    Ours                     & \multicolumn{1}{|c|}{\textbf{97.1±1.8}} & \textbf{98.3±1.2} & \multicolumn{1}{c|}{\textbf{92.4±4.3}} & \textbf{94.9±3.7} \\ \hline
    \end{tabular}
    }
    \label{table3}
\end{table}

\subsection{Few-shot Learning}\label{sec53}
Few-shot learning aims to tackle new tasks containing limited labeled training examples using prior knowledge. 
\noindent
\textbf{Dataset:} Here we conduct experiments on ModelNet40 \cite{ref24} to evaluate our method. All our experiments follow K-way, m-shot setting \cite{ref54}. Specifically, we randomly select K classes and sample m+20 samples for each class. The model is trained on K*m samples (support set), and evaluated on K*20 samples (query set). We compare our method with other competitors under “5 way, 10 shot”, “5 way, 20 shot”, “10 way, 10 shot”, “10 way, 20 shot” settings and report the mean and standard deviation over 10 runs. 

\noindent
\textbf{Competitors:} Our competitors can be roughly divided into three categories: (1) unsupervised methods; (2) supervised methods; (3) self-supervised pre-training methods. For the unsupervised methods \cite{ref39,ref56,ref57,ref58}, we train a linear SVM based on their extracted unsupervised representations. For supervised methods \cite{ref25,ref26,ref27,ref48,ref59} including our baseline, we train the model from scratch. For self-supervised pre-training methods \cite{ref12,ref37} including our McP-BERT, we fine-tune the model with pre-trained weight as initializations. 

\noindent
\textbf{Results:} The results are shown in Table \ref{table3}. 
We can see that when the labeled training data is insufficient, our method can still perform well. Our method has the highest average accuracy under four different settings. It's noticeable that compared with other methods, our method commonly has a smaller standard deviation, which indicates our method is more stable. Our method also significantly improves the baseline by 9.3\%, 5.2\%, 7.8\%, and 5.5\%, which demonstrates the strong generalization ability of our method.

\begin{table*}[]
    \centering
    \caption{
        \textbf{Part segmentation results on the ShapeNetPart} \cite{ref55}. 
        We report the mean $IoU$ across all part categories $mIoU_C$ (\%) and the mean $IoU$ across all instances $mIoU_I$ (\%), as well as the $IoU$ (\%) for each categories. * represents training from scratch.
    }
    \setlength{\tabcolsep}{4.6mm}{
    \begin{tabular}{l|c|c|c|c|c|c|c}
    \hline
    & PointNet  & PointNet++  & DGCNN & Transformer*    & OcCo  & Point-BERT  & Ours  \\ \hline
    $mIoU_C$      & 80.4              & 81.9                & 82.3           & 83.4          & 83.4          & 84.1                & \textbf{84.4} \\ 
    $mIoU_I$     & 83.7              & 85.1                & 85.2           & 85.1          & 85.1          & 85.6                & \textbf{86.1} \\ \hline
    \end{tabular}
    }
    \label{table4}
\end{table*}

\subsection{Point Cloud Part Segmentation}\label{sec54}
Point cloud part segmentation is a challenging task aimed at predicting point-wise label for point cloud. 
Here, we evaluate our method on the widely used ShapeNetPart \cite{ref55} dataset.

\noindent
\textbf{Dataset:} ShapeNetPart contains 16881 CAD models from 16 object categories, annotated with 50 parts. 
We follow the setting in \cite{ref25} to randomly sample 2048 points from each CAD model for training and test. 
We also double the number of local patches to 128 in the pre-training for part segmentation. 

\noindent
\textbf{Competitors:} We compare our method with some widely used methods \cite{ref25,ref26,ref27} and recently published self-supervised pre-training methods \cite{ref12,ref37}. 
We also adopt a standard transformer trained from scratch as our baseline.

\noindent
\textbf{Results:} We evaluate all the methods on two types of $mIoU$, i.e. $mIoU_C$ and $mIoU_I$. The former denotes the mean $IoU$ across all part categories and the later denotes the mean $IoU$ across all instances. As shown in Table \ref{table4}, our method achieves the best performance on both metrics. Our method also boosts baseline's performance while OcCo fails to do so. In addition, please refer to the supplementary material for the IOU of each category and the visualization of segmentation.

\subsection{Ablation Study}\label{sec55}
In this section, we conduct extensive experiments on ModelNet40 \cite{ref24} to study the effect of hyperparameters. 

\noindent
\textbf{The temperature coefficient $\tau$:}
The hyperparameter $\tau$ controls the smoothness of supervision signals. 
When $\tau$ is small, we will get a sharp probability distribution of token ids. 
And conversely, when $\tau$ is large, the probability distribution tends to be a uniform distribution. 
To study the effect of $\tau$, we conduct an ablation study on $\tau$ and the result is shown on Table \ref{table5}. 
We also add a single-choice version  i.e., the same as in Point-BERT as competitor, which equals to $\tau \rightarrow 0$. 
Observed from the result, we find our multi-choice strategy performs best when temperature coefficient is set to 0.005 empirically and improves previous BERT-style pre-training.

\begin{table}[]
    \centering
    \caption{
            \textbf{Ablation study on the temperature coefficient} $\tau$. 
            The ablation is conducted on point cloud classification downstream task on ModelNet40 \cite{ref24}.
    }
    {
    \begin{tabular}{l|c|cccc}
    \hline
    \multirow{2}{*}{}              & \multirow{2}{*}{\begin{tabular}[c]{@{}c@{}}Single-choice\\ (Point-BERT)\end{tabular}} & \multicolumn{4}{c}{Multi-choice}                                                           \\ \cline{3-6} 
                                   &                & \multicolumn{1}{c|}{0.005}         & \multicolumn{1}{c|}{0.05} & \multicolumn{1}{c|}{0.5}  & 5    \\ \hline
    \multicolumn{1}{c|}{Acc (\%)} & 93.8  & \multicolumn{1}{c|}{\textbf{94.1}} & \multicolumn{1}{c|}{93.6} & \multicolumn{1}{c|}{93.5} & 93.6 \\ \hline
    \end{tabular}
    }
    \label{table5}
\end{table}

\noindent
\textbf{The weight coefficient $\omega$:}
The hyperparameter $\omega$ balances the low-level semantics from tokenizer and high-level semantics from transformer encoder. 
We conduct an ablation study on $\omega$ and the result is shown on Table \ref{table6}. 
In this ablation study, we not only compare our McP-BERT under different $\omega$ settings, but also compare with a baseline, i.e., Point-BERT and a version without warm up strategy for $\omega$. 
In the version without warm up, we set $\omega=0.8$ at the beginning of the pre-training rather than progressively decreasing $\omega$ after pre-training several epochs. 
Observed from the result, the version without warm up collapses during the pre-training, because the transformer is not well-trained enough to provides accurate semantics. 
Moreover, pre-training seems to perform better as the weight coefficient $\omega$ goes larger. 
But when $\omega = 1.0$, the supervision signals don't incorporate high-level semantics, making the pre-training has the same performance as the baseline. 
As can be seen from the above, both the warm up strategy and the high-level semantics generated from transformer play important roles. See supplementary materials for more ablation studies.

\begin{table}[]
    \centering
    \caption{
            \textbf{Ablation study on the weight coefficient } $\omega$. 
            The ablation is conducted on point cloud classification downstream task on ModelNet40 \cite{ref24}.
    }
    \setlength{\tabcolsep}{2.5mm}
    {
    \begin{tabular}{l|c|c|c|c|c|c}
    \hline
                                   & 0    & 0.2  & 0.4  & 0.6  & 0.8  & 1.0 \\ \hline
    \multicolumn{1}{c|}{Acc (\%)} & 93.5 & 93.4 & 93.6 & 93.6 & \textbf{94.1} & 93.8     \\ \hline
    \multicolumn{1}{c|}{}& \multicolumn{3}{c|}{w/o warm up} & \multicolumn{3}{c}{Point-BERT} \\ \hline
    \multicolumn{1}{c|}{Acc (\%)}  & \multicolumn{3}{c|}{collapse} & \multicolumn{3}{c}{93.8} \\ \hline

    \end{tabular}
    }
    \label{table6}
\end{table}

\subsection{Visualization}\label{sec56}
\begin{figure}[h]
    \centering
    \includegraphics[width=0.5\textwidth]{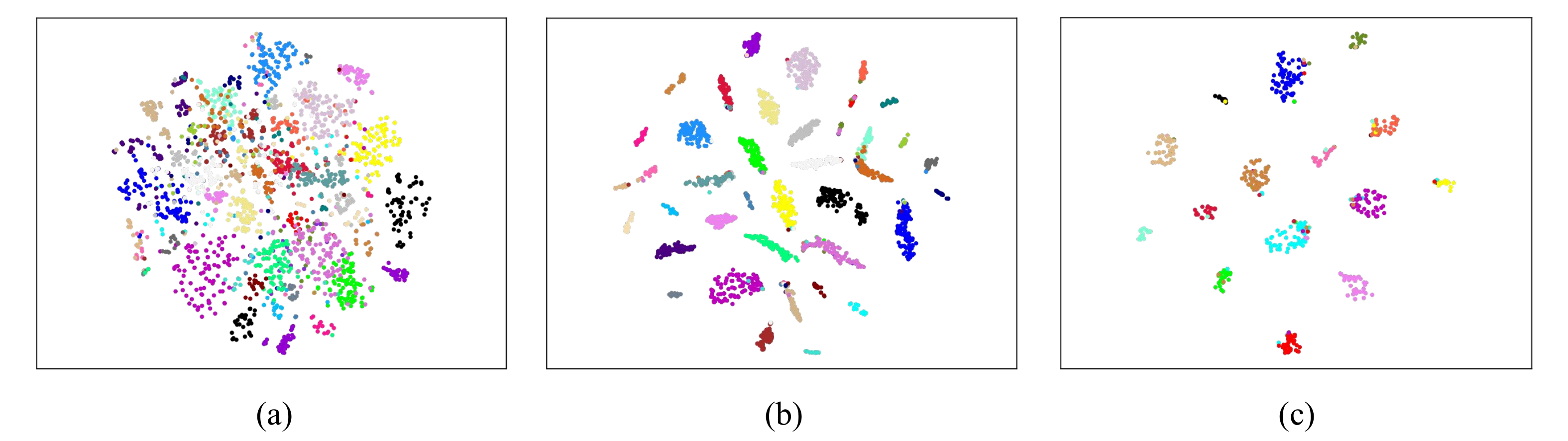}
    \caption{
        \textbf{Visualization of feature distributions.} 
        We utilize t-SNE \cite{ref60} to visualize the features learned by our McP-BERT. Features from different categories are visualized in different colors. (a) Features after pre-training; (b) Features fine-tuned on ModelNet40; (c) Features fine-tuned on ScanObjectNN. 
    }
    \label{fig5}
\end{figure}

To further understand the effectiveness of our method, we visualize the learned features via t-SNE \cite{ref60}. Figure \ref{fig5}(a) shows our learned features before fine-tuning. As observed from it, features are well separated even trained without annotations, which are suitable for model initialization. Figure \ref{fig5}(b) and (c) provide visualization of features fine-tuned on ModelNet40 \cite{ref24} and ScanObjectNN \cite{ref47}. As can be seen, features form multiply clusters that are far away from each other, indicating the effectiveness of our method.

\subsection{Computational Overhead}\label{sec57}
To demonstrate that our method improves the performance while introducing almost no extra computational overhead, we compare to the computational overhead of Point-BERT \cite{ref12} on the same device. 
The pre-training is implemented on an Intel Xeon Platinum 8260 CPU and two RTX3090 GPU. 
We select the time overhead of each pre-training epoch as our evaluation metric. 
As shown in Table \ref{table8}, our method only incurs 1\% extra time consuming, which is almost negligible.

\begin{table}[]
    \centering
    \caption{
       \textbf{Comparisons of the time overhead of pre-training.}
    }
    \setlength{\tabcolsep}{3.9mm}
    {
            \begin{tabular}{c|c|c}
            \hline
                                   & Point-BERT & Ours   \\ \hline
            Running time (s/epoch) & 143.48              & 144.94 \\ \hline
            \end{tabular}
    }
    \label{table8}
\end{table}

\section{Conclusion}
In this paper, we propose McP-BERT, a BERT-style pre-training method for point cloud pre-training, which tackles the problem of improper tokenizer in previous work. 
We release the previous strict single-choice constraint on token ids and utilize the probability distribution vector of token ids as supervision signals to avoid semantically-similar patches corresponding to different token ids. 
In addition, we use the high-level semantics generated by transformer to re-weight the probability distribution vector to further avoid semantically-dissimilar patches corresponding to the same token ids. 
Extensive experiments on different datasets and downstream tasks are conducted to evaluate the performance of our McP-BERT. 
The results show that our McP-BERT not only improves the performance of previous work on all downstream tasks with almost no extra computational overhead, but also achieve new state-of-the-art on point cloud classification and point cloud few-shot learning. 
The experiments also reveal that our pre-training method successfully transfer knowledge learned from unlabeled data to downstream tasks, which has great potential in point cloud learning.

\bibliography{ref}

\end{document}